\documentclass[oneside]{article}

\usepackage{PRIMEarxiv}
\usepackage[utf8]{inputenc} 
\usepackage[T1]{fontenc}    
\usepackage{xurl}
\usepackage{hyperref}       
\usepackage{url}            
\usepackage{booktabs}       
\usepackage{amsfonts}       
\usepackage{nicefrac}       
\usepackage{microtype}      
\usepackage{lipsum}
\usepackage{fancyhdr}       
\usepackage[dvipsnames]{xcolor}
\usepackage{graphicx}
\usepackage{comment} 
\graphicspath{{media/}}     
\usepackage[most]{tcolorbox}
\tcbuselibrary{breakable}
\usepackage{multicol}
\usepackage{hyperref}
\usepackage{array}      
\usepackage{booktabs}   
\usepackage[preprint]{neurips_2025}
\usepackage{multirow}
\usepackage{float}
\usepackage{pdflscape}
\usepackage{graphicx}
\usepackage{array}
\usepackage{enumitem}
\usepackage{algorithm}
\usepackage{algpseudocode}

\usepackage{booktabs}
\usepackage{pdflscape}    
\usepackage{geometry}     

\usepackage{lipsum}
\usepackage{rotating}
\usepackage{pifont}
\usepackage{hyperref}

\tcbset{
  promptbox/.style={
    colback=gray!10,                             
    colframe=green!50!black,                      
    sharp corners,                               
    boxrule=1pt,                                 
    left=2mm, right=2mm, top=1mm, bottom=1mm,    
    fontupper=\ttfamily,         
    fonttitle=\bfseries,                        
    title=Prompt,                                
    breakable                                   
  }
}
\newtcolorbox{promptbox}[1][]                    
  {promptbox,#1}

\tcbset{
  userchat/.style={
    enhanced,
    breakable,                         
    sharp corners,
    boxrule=1pt,
    colback=blue!5!white,              
    colframe=blue!70!black,            
    fonttitle=\bfseries\ttfamily,      
    fontupper=\ttfamily,               
    title=User,                        
    left=2mm, right=2mm, top=1mm, bottom=1mm,
  },
  aiassistant/.style={
    enhanced,
    breakable,
    sharp corners,
    boxrule=1pt,
    colback=green!5!white,             
    colframe=green!60!black,           
    fonttitle=\bfseries\ttfamily,
    fontupper=\ttfamily,               
    title=Assistant,                   
    left=2mm, right=2mm, top=1mm, bottom=1mm,
  }
}

\newtcolorbox{user}[1][]{userchat,#1}
\newtcolorbox{assistant}[1][]{aiassistant,#1}

\newcommand{\cmark}{\textcolor{green}{\ding{51}}}  
\newcommand{\xmark}{\textcolor{red}{\ding{55}}}    

\usepackage{changepage} 
\usepackage{tcolorbox}
\usepackage{lmodern}    

\usepackage{subcaption}

\newcommand{\mynote}[2]{\fbox{\bfseries\sffamily\scriptsize{#1}}
	{\small\textsf{\emph{#2}}}}
\newcommand{\cc}[1]{{\leavevmode\color{orange}\mynote{Cristian:}{#1}}}

\definecolor{codegreen}{rgb}{0,0.6,0}
\definecolor{codegray}{rgb}{0.4,0.4,0.4}
\definecolor{codepurple}{rgb}{0.32,0,0.72}
\definecolor{backcolour}{rgb}{0.95,0.95,0.95}

\usepackage{xcolor}
\usepackage{soul}
\definecolor{lightgreen}{rgb}{0.6, 0.87, 0.54}  
\definecolor{lightorange}{rgb}{1.0, 0.73, 0.47} 
\definecolor{lightpurple}{rgb}{0.77, 0.69, 0.84} 
\newcommand{\hlightgreen}[1]{\sethlcolor{lightgreen}\hl{#1}}
\newcommand{\hlightorange}[1]{\sethlcolor{lightorange}\hl{#1}}
\newcommand{\hlightpurple}[1]{\sethlcolor{lightpurple}\hl{#1}}

\title{HealthBranches: Synthesizing Clinically-Grounded Question Answering Datasets via Decision Pathways}

\author{
  Cristian Cosentino\thanks{Equal contribution.} \\
  University of Calabria\\
  Rende, Italy\\
  \texttt{cristian.cosentino@dimes.unical.it} \\
  \And
  Annamaria Defilippo\footnotemark[1]\\
  University of Catanzaro\\
  Catanzaro, Italy\\
  \texttt{annamaria.defilippo@unicz.it} \\
    \And
  Marco Dossena\footnotemark[1]\\
  University of Eastern Piedmont\\
  Alessandria, Italy\\
  \texttt{marco.dossena@uniupo.it} \\
  \And
  Christopher Irwin\footnotemark[1]\\
  University of Eastern Piedmont\\
  Alessandria, Italy\\
  \texttt{christopher.irwin@uniupo.it} \\
  \And
  Sara Joubbi\footnotemark[1]\\
  University of Pisa\\
  Pisa, Italy\\
  \texttt{sara.joubbi@phd.unipi.it} \\
  \And
  Pietro Liò\\
  University of Cambridge\\
  Cambridge, England\\
  \texttt{pl219@cam.ac.uk}\\
  }

\begin{document}
\maketitle

\begin{abstract}
HealthBranches is a novel benchmark dataset for medical Question-Answering (Q\&A), specifically designed to evaluate complex reasoning in Large Language Models (LLMs). This dataset is generated through a semi-automated pipeline that transforms explicit decision pathways from medical source into realistic patient cases with associated questions and answers. Covering 4,063 case studies across 17 healthcare topics, each data point is based on clinically validated reasoning chains. HealthBranches supports both open-ended and multiple-choice question formats and uniquely includes the full reasoning path for each Q\&A. Its structured design enables robust evaluation of LLMs’ multi-step inference capabilities, including their performance in structured Retrieval-Augmented Generation (RAG) contexts. HealthBranches establishes a foundation for the development of more trustworthy, interpretable, and clinically reliable LLMs in high-stakes domains while also serving as a valuable resource for educational purposes.
\end{abstract}


\section{Introduction}
Large Language Models (LLMs) have emerged as powerful tools for generating and analyzing human-like text, leveraging extensive corpora and transformer-based architectures \cite{achiam2023gpt}. They have enabled significant advances in Natural Language Processing (NLP) tasks such as translation, summarization, information retrieval, and question-answering (Q\&A) \cite{naveed2023comprehensive, pan2306unifying}. 

Q\&A, in particular, has greatly benefited from LLMs, with models like ChatGPT supporting public health by providing information on diseases, prevention, and medical services \cite{biswas2023role}.

However, despite their promise, the deployment of LLMs in medical and clinical contexts presents significant challenges. These include limited accuracy, inherent biases, data constraints, lack of contextual understanding, and hallucinations, where the model generates plausible but incorrect information \cite{xiong2024benchmarking, arslan2024survey}. 
In medical applications, such inaccuracies pose serious risks, as misleading information can directly impact patient health and violate privacy regulations \cite{shi2024medical, carlini2021extracting}. Additionally, LLM training corpora may lack up-to-date medical knowledge or fail to provide contextually relevant insights, further limiting their reliability in healthcare settings \cite{xiong2024benchmarking}. 

To address these limitations, three main strategies have been proposed: fine-tuning on domain-specific data \cite{rafailov2024direct}, Retrieval-Augmented Generation (RAG) \cite{lewis2020retrieval}, and prompt engineering \cite{liu2023pre}.

RAG has gained particular attention for enhancing LLM performance by retrieving external domain knowledge \cite{lewis2020retrieval, guu2020retrieval, izacard2023atlas, huo2023retrieving}, reducing hallucinations and improving factual accuracy \cite{gao2023retrieval}. Integrating RAG with LLMs has notably improved performance in medical Q\&A by enhancing reasoning capabilities. In this context, LLMs provide diagnosis and personalized advice \cite{jin2024health,wang2024healthq}, while in domain-specific Q\&A, they offer specialized insights \cite{gilson2024enhancing}.

However, in medical applications, the precision of the retrieved content is critical. Knowledge Graphs (KGs) help mitigate noise by structuring domain-specific knowledge in interpretable and verifiable ways \cite{procko2024graph, jia2024medikal}. 
This is achieved through approaches such as graph-based RAG, which leverages KGs for interconnected reasoning \cite{wu2024medical}; task-specific KG generation, enabling dynamic adaptation to emerging medical knowledge \cite{procko2024graph}; and structured prompts with KG interaction, which guide models to reduce hallucinations and ensure evidence-based responses \cite{kharitonova2024leveraging}.

Despite these advances, there is still a lack of datasets and tools that closely integrate structured medical knowledge with question-answering tasks, reflecting real-world diagnostic complexity and supporting RAG pipelines.

In this study, we introduce HealthBranches, a new medical Q\&A dataset that combines structured reasoning paths, various question formats (including open-ended and multiple-choice), and rich semantic annotations. Built from realistic patient scenarios grounded in validated medical knowledge, HealthBranches supports interpretable and clinically meaningful evaluation of LLMs. Each question includes an explicit reasoning path, enabling detailed assessments of a model’s step-by-step reasoning, beyond final answer accuracy. To construct the dataset, we developed a semi-automated generation pipeline that extracts knowledge graphs from medical sources to synthesize patient cases and diagnostic reasoning aligned with real-world clinical practices. The resulting questions were validated through LLM-assisted refinement, automated evaluation, and human-in-the-loop auditing. We evaluated several LLMs under zero-shot and RAG settings, demonstrating that HealthBranches serves as a valuable benchmark for assessing reasoning capabilities in medical Q\&A.

\section{Related Work}

Training LLMs in the medical domain requires massive datasets, including extensive text corpora derived from Electronic Health Records (EHRs) and medical books. There are several Q\&A benchmarks on different medical topics, with different types of data and models tested on them, shown in Table \ref{tab:datasets}. 
In particular, closed-ended multiple-choice Q\&A datasets, characterized by fixed-answer formats with a limited set of options per question, are designed to address tasks in medical dialogue and biomedical text analysis \cite{vilares2019head,suri2021mediaqa,jin2021disease,jin2019pubmedqa}. In contrast, open-ended Q\&A datasets focus on generating free-form responses and often require manual verification to ensure accuracy \cite{khandekar2024medcalc,he2020pathvqa}. Multi-task and multi-modal benchmarks have been developed to assess diverse tasks such as text generation, reasoning, and domain-specific understanding; these benchmarks integrate both textual and visual data and are evaluated under zero-shot or few-shot conditions \cite{hendrycks2020measuring,zuo2025medxpertqa}. Additionally, structured query and dialogue-based datasets facilitate the processing of complex queries on structured data, such as EHRs, and capture real-world interactions in medical conversations, such as in \cite{park2021knowledge}. 
Notably, \cite{zhang2024ultramedical} presents a collection of high-quality manual and synthetic datasets in the field of biomedicine, significantly advancing biomedical LLM fine-tuning, while \cite{sivasubramaniam2024sm3} introduces SM3-Text-to-Query, the first multi-model medical Text-to-Query benchmark based on synthetic patient data following the SNOMED-CT taxonomy, enabling evaluations across various query languages and database models.

Compared to existing benchmarks, HealthBranches uniquely combines multiple-choice and open-ended formats with explicit reasoning paths derived from medical knowledge graphs. As shown in Table~\ref{tab:datasets}, few prior datasets support both qualitative reasoning and step-by-step explanations without relying on computational tasks. While MedQA and PubMedQA assess medical knowledge, they lack explanation structure. MedCalc-BENCH provides explanations but focuses on numeric computation. In contrast, HealthBranches enables the evaluation of non-computational, interpretable, and knowledge-driven reasoning, making it well-suited for training and evaluating trustworthy medical LLMs. MedCalc-BENCH provides detailed, verified explanations but focuses on quantitative clinical calculations, using formula-based reasoning. In contrast, HealthBranches enables the evaluation of non-computational, interpretable, and knowledge-driven reasoning. This makes it particularly well-suited for training and evaluating LLMs on tasks that require structured, qualitative clinical decision-making rather than numeric computation.

\begin{table}[!ht]
  \centering
  \resizebox{\linewidth}{!}{%
    \begin{tabular}{@{} l r l c c c c @{}}
      \toprule
      \textbf{Dataset (Year)}  & \textbf{\# questions} & \textbf{Type of questions}       & \textbf{Knowledge} & \textbf{Qual.\ Reasoning} & \textbf{No-Comput.} & \textbf{Explanation} \\
      \midrule
      HeadQA (2019) \cite{vilares2019head}            &  6,765  & Multiple choice                & \cmark & \cmark & \cmark & \xmark \\ 
      MEDCALC-BENCH (2024) \cite{khandekar2024medcalc} &  1,000  & Open-ended                     & \cmark & \cmark & \xmark & \cmark \\ 
      MeDiaQA (2021) \cite{suri2021mediaqa}            & 23,048  & Multiple choice                & \cmark & \xmark & \cmark & \xmark \\ 
      MedQA (2020) \cite{jin2021disease}               & 61,097  & Multiple choice                & \cmark & \cmark & \cmark & \xmark \\ 
      MedXpertQA (2025) \cite{zuo2025medxpertqa}       &  4,460  & Multi-modal questions          & \cmark & \cmark & \cmark & \xmark \\ 
      MIMIC-SPARQL (2020) \cite{park2021knowledge}     & 10,000  & Table- and graph-based queries & \cmark & \xmark & \xmark & \xmark \\ 
      PathVQA (2020) \cite{he2020pathvqa}              & 32,799  & Open-ended                     & \cmark & \xmark & \cmark & \xmark \\ 
      PubMedQA (2019) \cite{jin2019pubmedqa}           &273,500  & Multiple choice                & \cmark & \cmark & \cmark & \xmark \\ 
      \hline
      \textbf{HealthBranches} (2025)                   &  4,063  & Multiple choice \& open-ended  & \cmark & \cmark & \cmark & \cmark \\ 
      \bottomrule
    \end{tabular}%
  }
  \caption{Summary of existing medical Q\&A datasets and comparison with \textbf{HealthBranches}. We assess each dataset along four qualitative dimensions:
(1) \textbf{Knowledge} – whether the dataset tests knowledge to a particular domain;
(2) \textbf{Qualitative Reasoning} – whether the dataset tests logical or conceptual reasoning rather than quantitative computation;
(3) \textbf{No Comput.} – whether the dataset does not require quantitative calculations, formulas, or numeric estimation;
(4) \textbf{Explanation} – whether the dataset includes a step-by-step justification or reasoning trace.
While these criteria are inherently qualitative, we follow prior precedent in MedCalc-BENCH \cite{khandekar2024medcalc}. \cmark = present, \xmark = absent.}
  \label{tab:datasets}
\end{table}

\section{Proposed Methodology}
\label{proposed_metodology}
The proposed methodology is aimed at improving the reliability of LLMs in highly specialized domains. We define a structured framework capable of grounding Q\&A generation in domain-specific knowledge, ensuring factual consistency and interpretability, leveraging structured medical knowledge sources. 

\begin{figure}[ht]
    \centering
    \includegraphics[width=1.0\linewidth]{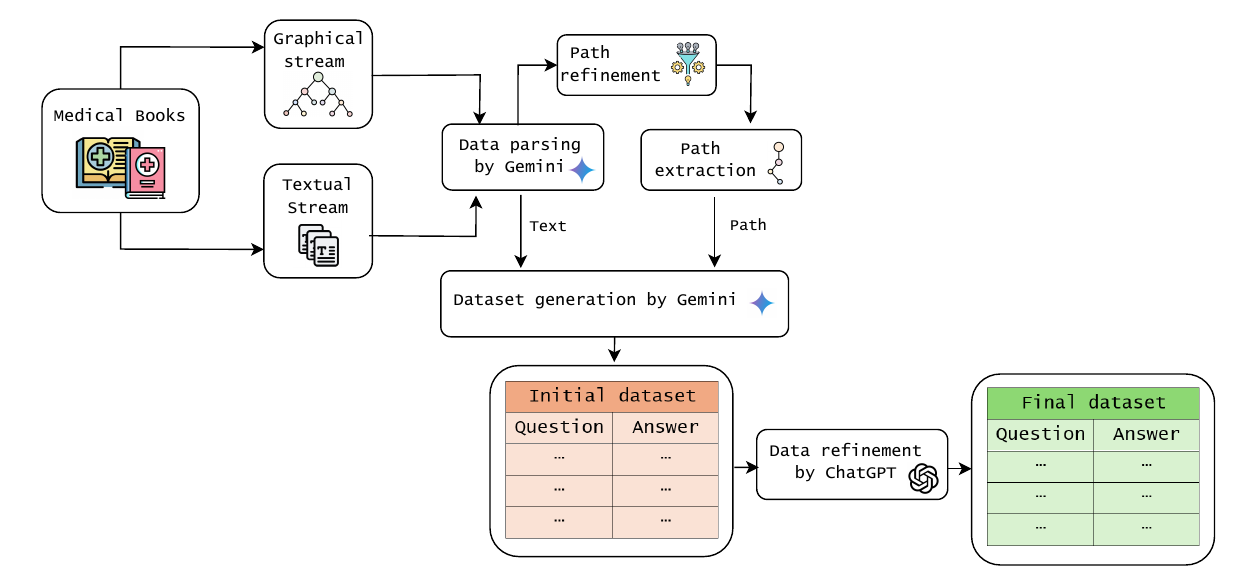}
    \caption{
    Workflow for dataset construction illustrating the sequential stages of data extraction, transformation, and Q\&A pair generation. 
    \textit{Parsing the knowledge source}: textual and graphical streams are extracted from medical books and parsed. \textit{Path extraction and refinement}: graph-based representations are processed to extract and refine relevant semantic paths. \textit{Q\&A generation}: an LLM is prompted with textual and path information to generate initial question–answer pairs. \textit{Q\&A refinement}: the dataset is further refined using another LLM to ensure consistency and quality.}
    \label{fig:dataset_generation}
\end{figure}

This section introduces the dataset and the domain context, followed by a detailed description of the construction pipeline (as shown in Figure \ref{fig:dataset_generation}), which encompasses knowledge parsing, Q\&A generation, and refinement. Finally, we outline the dataset properties and the evaluation setup employed to assess LLM performance.

\subsection{Dataset}
\label{sec:dataset}
We propose an automatic pipeline, augmented by a Human-in-the-Loop component that only checks and approves a subset of the generated questions, for building medical question-answering datasets from structured knowledge sources, addressing challenges such as data scarcity and privacy concerns.
The method leverages medical reference materials (e.g., textbooks) containing textual descriptions of patient treatment protocols and decision graphs that formalize clinical decision-making processes.
By sampling realistic decision paths, the pipeline generates plausible patient scenarios, as a foundation for meaningful Q\&A grounded in authoritative medical knowledge.

The resulting dataset and the pipeline offer valuable applications: it can be used to create educational resources for medical students, evaluate the proficiency of current large language models (LLMs) in medical reasoning tasks and help building more clinical grounded decision support systems.

An overview of the followed pipeline is shown in Figure~\ref{fig:dataset_generation}, while implementation details and the prompt are provided in Section~\ref{app:dataset_details} of the Supplementary Material.
Figure~\ref{fig:qea_example} illustrates an example: from a clinical path on dyspnea, a realistic scenario is generated, followed by a multiple-choice question. The correct answer is derived directly from the structured reasoning, demonstrating the pipeline’s effectiveness in producing reliable medical Q\&A data.

\begin{figure}[ht]
    \centering
    \includegraphics[width=1\linewidth]{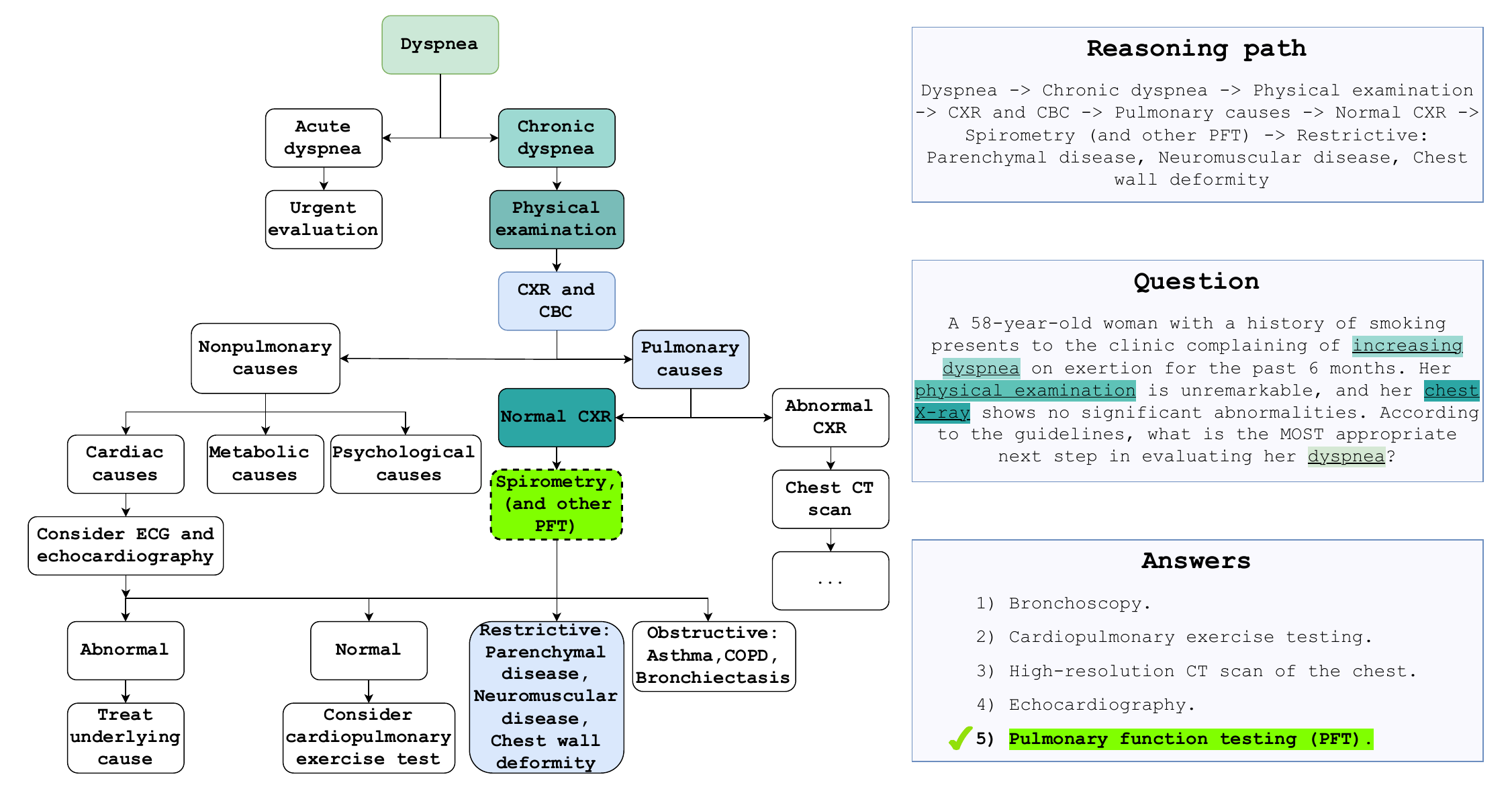}
    \caption{Detailed example of Q\&A generation process, from path extraction through Q\&A pair construction. On the left side an example of an extracted decision graph that presents the steps to treat dyspnea. On the right the extracted reasoning path and the generated question-answer pair.}
    \label{fig:qea_example}
\end{figure}

\textbf{Parsing of the Knowledge Source} The knowledge source consists of medical textbooks \cite{book1, book2, book3} divided into sections, each detailing the management of a patient presenting a specific symptom or condition via a clinically validated algorithmic framework. Here, “algorithmic framework” refers to a graph‐based representation (most commonly a decision tree) in which each node corresponds to a diagnostic or therapeutic step (see Figure \ref{fig:qea_example}). This structure systematically encapsulates the clinical reasoning and domain‐specific knowledge concerning to the symptom or condition referred to one of the 17 clinical domains. Moreover, the applied pipeline is applicable to other medical domains, making it generalizable beyond those covered in the study.

Consequently, every section comprises two parallel streams of information: the \textit{Textual stream} is a continuous narrative describing the clinical rationale and procedural steps. The \textit{Graphical stream} is a knowledge graph (decision tree) that imposes structure on the reasoning process.
Both streams are extracted automatically using Gemini-flash 2.0 \cite{gemini}, yielding (a) plain‐text files conveying the text content and (b) a files encoding the graph structure. 

\textbf{Path Extraction and Refinement} From each graph, we enumerate all root-to-leaf traversals, retaining up to two distinct paths per leaf to ensure a balance between coverage and dataset size. Subsequently, we refine each path using Gemini-flash 2.0 (refer to Supplementary Material \ref{app:dataset_details}), normalizing terminology and removing formatting artifacts while maintaining clinical semantics.

\textbf{Q\&A Generation} The dataset’s Q\&A content was generated by prompting Gemini‑flash 2.0 to formulate relevant questions based on the text associated with a specific condition and a selected reasoning path. First, Gemini was instructed to produce an open‑ended answer using all contextual information from the patient scenario derived via the reasoning path allowing a fully detailed, free‑form response. Second, that open‑ended answer was reused as the correct option in a multiple‑choice (quiz) version of the item.
Furthermore, we requested that Gemini produce four incorrect answers for each question Notably, the prompt encouraged generating distractors that were plausible within the context of the specific condition yet not valid within the provided reasoning path. This approach yielded incorrect options which are non-obvious in general, and more challenging than randomly generated distractors. The result is a dataset where each reasoning path is linked to a question-answer pair, along with a set of incorrect answers, which can be used for benchmarking downstream models in either the open-question or quiz versions of the dataset.

\textbf{Q\&A Refinement} The generation process described above may occasionally yield flawed examples, leading to unanswerable questions, correct answers that contain inconsistencies, errors, or minor details that misalign with clinical guidelines. To investigate the effects of model scale on question difficulty and robustness, we incorporate Llama 3.1 (405 B) \cite{llama3} only in the refinement phase, as it significantly surpasses the capacity of our existing benchmark models. This integration aims to identify intrinsically flawed questions that remain incorrect regardless of the model's parameter count. To address these issues, we implemented a two-stage refinement procedure. First, we identified all generated questions that were incorrectly answered by both Llama3.3 (70 B) and Llama3.1 (405 B) \cite{llama3}. From this subset, we retained only the questions that were misanswered by at least half of the models in our benchmark (i.e. 5 models, excluding Llama3.3 (70 B)), resulting in a pool of 1,203 questions for further examination.
Each question was submitted to GPT-4o with web search and reasoning capabilities enabled and subsequently refined according to a defined protocol. Following the audit by GPT-4o, each question was also evaluated by a human reviewer. The results were compared, leading to a final decision on whether to modify the question, adjust or change the correct option, or eliminate the question if it was found to be fundamentally flawed. Detailed information on the process, including flawed examples, the protocol, and the GPT-4o prompt, is provided in Appendix \ref{app:qa_valid}.

The main motivation for hybridizing ChatGPT and Gemini was to reduce potential systematic biases from relying on a single model throughout the pipeline stages: generation, flawed path identification, and refinement. This diversity mitigates the risk of model-specific artifacts and enhances robustness and generalizability.

\subsection{Evaluation}

The evaluation strategies employed to assess the proficiency of publicly available LLMs were applied to both the quiz and the open-ended answers version of the dataset. In both dataset versions, the models were evaluated in topline and benchmark settings: \textbf{Topline}: Models receive (a) only the reasoning path, (b) only the textual description, or (c) both as context. This isolates the value of structured reasoning support. \textbf{Benchmark}: Models answer with the standard zero‑shot Q\&A prompt (question only) with and without RAG context.
Regarding the RAG-based experiment, the implementation details are the followings:
\begin{itemize}
    \item Embedder Model: \texttt{mxbai-embed-large}
    \item Chunk Size: \texttt{500}
    \item Chunk Overlap: \texttt{150}
    \item Split Strategy: \texttt{RecursiveCharacterTextSplitter (LangChain)}
\end{itemize}

All the prompts used for evaluation are available in Appendix \ref{app:supp_quiz} (quiz) \ref{app:supp_open} (open-ended answer). The topline setting serves as a metric for evaluating the extent to which a model depends on the reasoning path to achieve a specific score. Additionally, it enhances the evaluation phase of the open-answer version of the dataset by providing a way to check that the reasoning path followed by the evaluated model is sound, rather than solely relying on the ground-truth answer.

\textbf{Models}
\label{models}
In our benchmark study on the HealthBranches dataset, we selected ten open, decoder-only LLMs that are freely available and deployable via the Ollama platform \cite{ollama}: Mistral 7B \cite{jiang2024identifying}, Llama 2 7B \cite{touvron2023llama}, Llama 3.1 8B \cite{grattafiori2024llama}, Gemma 7B \cite{team2024gemma}, Gemma 2 9B \cite{team2024gemma2}, Gemma 3 4B \cite{team2025gemma}, Qwen 2.5 7B \cite{yang2024qwen2}, DeepSeek-R1 8B \cite{guo2025deepseek}, Phi-4 14B \cite{abdin2024phi}, and Mistral NeMo 12B. To ensure a consistent evaluation framework, all models were evaluated in chat mode under zero-shot settings without chain-of-thought prompting.
\newline
In addition to the ten open models accessed locally via Ollama, we also evaluated three larger models: Llama 3.3 70B and Llama 3.1 405B \cite{grattafiori2024llama}(used only in the question control phase) through the Together AI platform \cite{toge}. This evaluation adhered to the same zero-shot chat configuration without chain-of-thought prompting. 

\textbf{Exact Match} In our quiz evaluation framework, we provided explicit instructions via the system prompt for models to respond with a single letter (A–E). To systematically handle variations (e.g., ``The correct answer is A''), we employed a regular expression pattern to extract only the answer letter from each response. After isolating the letter, we compared it to the correct answer key and calculated the overall accuracy as the proportion of extracted letters that matched the ground truth. The algorithm to extract the option from a string is described in Appendix \ref{app:rules}.

\textbf{LLM-as-a-judge Score} Ensuring the reliability and clinical coherence of generated responses is essential in medical question answering. To achieve this, our system incorporates an LLM-based Judge ({\textit{Gemini-flash 2.0}) that leverages in-context learning to provide a systematic, data-driven evaluation of each response. By leveraging the ground truth answer and the reasoning path, the judge assesses answers based on a predefined set of evaluation criteria, producing a score between 0 and 10 for each question-answer pair. Specifically, we employed the G-Eval metric presented by \cite{g-eval}. This approach operates in two distinct phases: in the first phase, the judge model is introduced to the task and provided with basic evaluation criteria to establish a set of evaluation steps. In the second phase, the generated steps are combined with the evaluation criteria and task introduction, enabling the model to return a score for each question-answer pair. The LLM-as-a-judge remains constant across evaluations. Detailed information regarding the implementation and the specific prompt used in the experiments can be found in Appendix \ref{app:supp_judge}.

\textbf{Semantic Similarity Score} We further assess the models by using a semantic similarity metric to compare the ground-truth answer with the open answer generated by each model under evaluation. To achieve this, we relied on the BGE-M3 model \cite{bge}, which is currently regarded as one of the most advanced publicly available models for text embedding. This model is capable of producing different types of embeddings (dense, sparse, ColBERT-style) and integrating them effectively. This versatility is particularly beneficial, as it allows to compare a pure semantic score given by the dense embeddings with a more lexical score obtained from the sparse embeddings.
The purpose of this second metric is twofold: first, it provides a validation check on the reliability of the LLM-as-a-judge metric; second, it offers insights into the specific strategies employed by models in addressing the open-ended question answering task.

\section{Experiments and Findings}
In the following sections, we highlight the main findings that emerged during the pipeline creation and the evaluation process of off-the-shelf open LLMs. In the first part, we focus on the dataset creation phase, outlining the impact of different refinements made in the pipeline and the results of the expert evaluation. The second part presents the results of evaluating different LLMs against the final datasets in both the quiz and open-ended answer settings.

\subsection{Dataset Characteristics}
After filtering, the final \textit{HealthBranches} dataset consists of 4,063 question–answer pairs across a wide range of medical specialties. The distribution of questions in these categories is shown in Table~\ref{tab:dataset_comp}, covering areas such as cardiology, neurology, infectious diseases, and hematology/oncology. This diversity ensures that the dataset reflects both common and specialized domains encountered in clinical practice.

As shown in Table~\ref{tab:datasets}, some macro-categories with a similar number of sub-conditions (e.g., \textit{Hematology/Oncology} and \textit{Rheumatology}) show significant variation in the number of generated questions. This discrepancy results from differences in the structure of the underlying clinical reasoning graphs. Categories with more complex conditions tend to support a greater variety of reasoning paths, leading to a higher number of distinct questions. A useful proxy for this complexity is the number of leaf nodes in the reasoning graph for each category, as each leaf node can contribute up to two sampled reasoning paths. Importantly, we chose not to artificially increase the number of questions for simpler categories to maintain semantic diversity and avoid redundancy in the dataset.

As highlighted in Table~\ref{tab:datasets}, \textit{HealthBranches} offers a unique combination of features compared to existing medical Q\&A benchmarks. It includes both multiple-choice and open-ended questions, supporting evaluation formats based on answer selection and free-text generation. Each question is aligned with a clear, structured reasoning path, similar to a clinical decision tree, that integrates domain-specific knowledge checks with qualitative reasoning steps and interpretable explanations. This structure enables detailed evaluation of reasoning ability, making \textit{HealthBranches} particularly well-suited for training and evaluating trustworthy, explanation-aware medical LLMs.

\begin{table}[ht]
\centering
\caption{Distribution of questions and number of leaf nodes across clinical categories in the HealthBranches dataset.}
\begin{tabular}{lrrr}
\toprule
\textbf{Category} & \textbf{Questions} & \textbf{Conditions} & \textbf{Leafs} \\
\midrule
Hematology/Oncology    & 531 & 22 & 147 \\
General Medicine       & 457 & 15 & 228 \\
Infectious Diseases    & 441 & 15 & 241 \\
Neurology               & 373 & 19 & 258 \\
Women's Health           & 350 & 18 & 152 \\
Gastrointestinal        & 338 & 21 & 239 \\
Nephrology              & 279 & 17 & 138 \\
Cardiology              & 278 & 18 & 153 \\
Rheumatology            & 217 & 22 & 197 \\
Behavioral Medicine    & 190 &  8 &  79 \\
Endocrinology           & 183 & 16 & 117 \\
Pharmacology            & 146 &  8 &  68 \\
Pulmonary Disease      & 125 & 12 &  97 \\
Emergency Medicine     &  54 &  7 &  36 \\
Urology                 &  45 &  5 &  30 \\
Dermatology             &  35 &  5 &  25 \\
Ocular                  &  21 &  4 &  22 \\
\bottomrule
\end{tabular}
\label{tab:dataset_comp}
\end{table}

\subsection{Experiments on Dataset Creation} 
\textbf{Q\&A validation} As discussed in Section~\ref{sec:dataset}, the automatic generation process may introduce flawed or imprecise examples. Figure~\ref{fig:refinement_diff_A} shows the accuracy gains after applying the refinement protocol, with clear improvements in both zero-shot and RAG-based settings particularly for larger models. Figure~\ref{fig:refinement_diff_B} summarizes the types and frequencies of corrections performed, such as adjusting clinical phrasing, fixing inconsistencies, or replacing incorrect answers.

\begin{figure}[ht]
  \centering
  \begin{subfigure}[t]{0.4\linewidth}
    \centering
    \includegraphics[width=\linewidth]{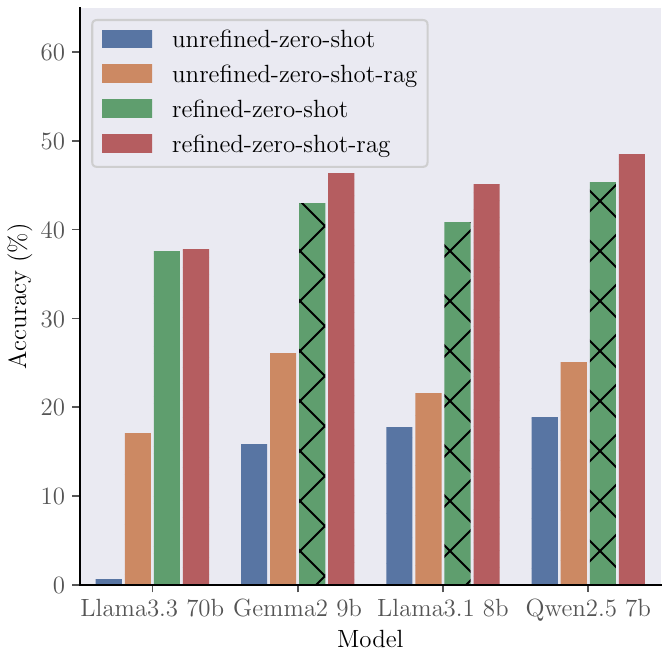}
    \caption{Accuracy improvement after Q\&A refinement}
    \label{fig:refinement_diff_A}
  \end{subfigure}
  \hfill
  \begin{subfigure}[t]{0.55\linewidth}
    \centering
    \includegraphics[width=\linewidth]
    {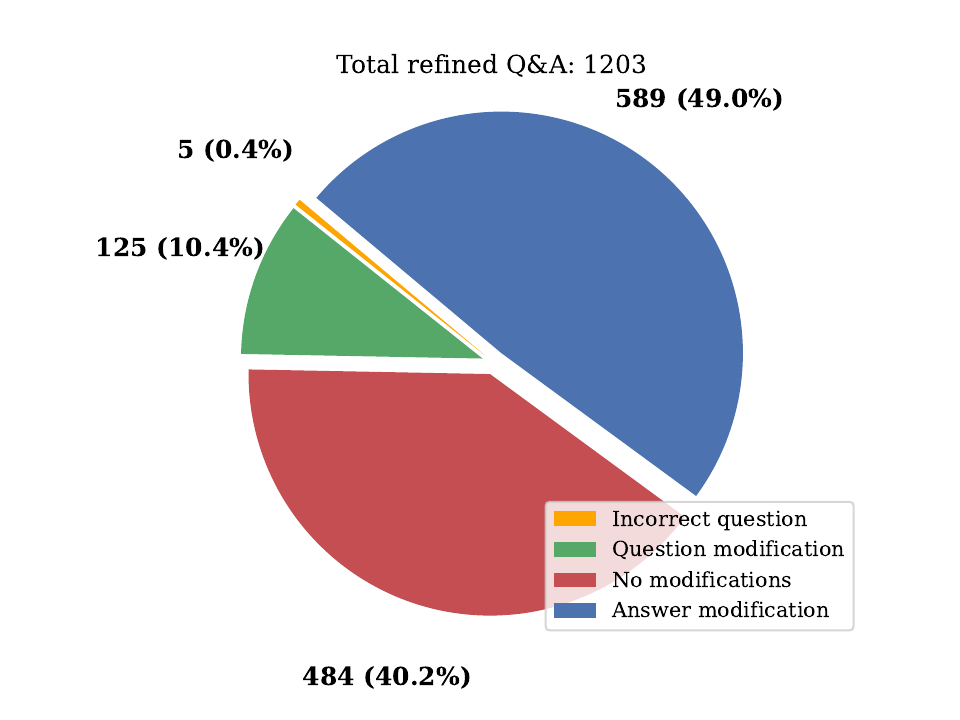}
    \caption{Distribution of refinement types}
    \label{fig:refinement_diff_B}
  \end{subfigure}

\caption{Impact of the Q\&A refinement process. Left: accuracy improvement across models. Right: type and frequency of applied refinements.}
  \label{fig:refinement_diff}
\end{figure}

\textbf{Expert Evaluations} We conducted an initial validation of the resulting dataset with a panel of practicing physicians and senior medical students, following the structured protocol in Appendix~\ref{app:med_val}. Each item was rated on the quality of the question content, the precision of the answers, and the reasoning (0 to 5 per aspect; total score of 15). The identities of the reviewers remain confidential, but the evaluation interface and aggregate statistics will be continuously updated and available on the dataset page. Preliminary results (Table~\ref{tab:reviewer_stats}) show that physicians and specialists gave more conservative ratings, while students tended to score higher. Only 19 questions received a score of 3 or below in at least one category, and 6 fell below 9/15 overall (refer to Appendix~\ref{app:med_spec} for some examples).
This ongoing validation suggests our synthetic pipeline produces clinically sound Q\&A content, although future work could enhance evaluation consistency by extending expert review across a broader set of quizzes and open questions.

\begin{table}[ht]
\centering
\caption{Reviewer performance and feedback statistics by expertise level for randomly selected questions.}
\label{tab:reviewer_stats}
\scriptsize 
\begin{tabular}{lcccccc}
\toprule
\textbf{Expertise} & \textbf{Question} & \textbf{Answer} & \textbf{Path} & \textbf{Mean} & \textbf{Reviewer} & \textbf{Reviews}\\
\midrule
Doctor/Physician   & 4.85/5 & 4.62/5 & 4.54/5 & 14.01/15 & 3  & 13  \\
Medical Specialist & 4.45/5 & 4.36/5 & 4.27/5 & 13.08/15 & 5  & 22  \\
Medical Student    & 4.74/5 & 4.79/5 & 4.72/5 & 14.25/15 & 15 & 102 \\
\midrule
\textbf{Total}     & 14.04/15 & 13.77/15 & 13.52/15 & 13.78/15 & 23  & 137 \\
\bottomrule
\end{tabular}
\end{table}

\subsection{Performance on the Generated Dataset}
The models were evaluated in different topline settings that focus on assessing how much the reasoning path impacts performance. Figure \ref{fig:bench_results}-top compares the results on the dataset using the quiz modality while Figure \ref{fig:bench_results}-middle/bottom does the same for the open-ended answer setting. Notably, it is clear that the models benefit significantly from having the reasoning path information, but this improvement is not observed in the topline with only textual information, which indicates that the models are not capable of extracting the reasoning paths using only the textual information.

\begin{table}[ht]
\centering
\caption{Model performance by modality. Each modality (Zero-shot, RAG, etc.) includes Accuracy, Judge score, and Similarity. Scores highlighted in \textcolor{red}{red} denote the best-performing model (rank 1), while those in \textcolor{blue}{blue} indicate the second-best performance (rank 2) for the corresponding metric.}
\scriptsize
\setlength{\tabcolsep}{3.5pt}
\resizebox{\textwidth}{!}{
\begin{tabular}{lc
  rrr  
  rrr  
  rrr  
  rrr  
  rrr  
}
\toprule
& \textbf{Model}
& \multicolumn{3}{c}{\textbf{Zero-Shot}} 
& \multicolumn{3}{c}{\textbf{Zero-Shot RAG}} 
& \multicolumn{3}{c}{\textbf{Topline Path}} 
& \multicolumn{3}{c}{\textbf{Topline Text}} 
& \multicolumn{3}{c}{\textbf{Topline All}} \\
\cmidrule(lr){3-5}
\cmidrule(lr){6-8}
\cmidrule(lr){9-11}
\cmidrule(lr){12-14}
\cmidrule(lr){15-17}
& 
& Acc & Judge & Sim
& Acc & Judge & Sim
& Acc & Judge & Sim
& Acc & Judge & Sim
& Acc & Judge & Sim \\
\midrule
Llama2 7b      &  & 19.71 & 5.14 & 0.314 & 34.36 & 4.89 & 0.317 & 28.25 & 8.15 & 0.385 & 40.09 & 4.82 & 0.320 & 55.45 & 6.68 & 0.360 \\
Mistral 7b     &  & 51.93 & 6.46 & 0.339 & 61.14 & 6.57 & 0.346 & 87.69 & 9.00 & 0.437 & 62.86 & 6.59 & 0.356 & 82.40 & 8.21 & 0.420 \\
Gemma 7b       &  & 57.27 & 5.24 & 0.305 & 62.22 & 5.73 & 0.325 & 89.37 & 8.67 & 0.407 & 61.90 & 6.08 & 0.340 & 80.56 & 8.23 & 0.407 \\
Deepseek-r1 8b &  & 61.33 & 5.46 & 0.321 & 63.97 & 6.02 & 0.337 & 88.48 & 8.41 & 0.418 & 60.94 & 6.16 & 0.334 & 76.15 & 7.43 & 0.377 \\
Gemma3 4b      &  & 62.79 & 6.14 & 0.318 & 62.79 & 6.20 & 0.320 & 89.74 & 8.91 & 0.391 & 61.92 & 6.06 & 0.330 & 79.30 & 8.18 & 0.383 \\
Llama3.1 8b    &  & 67.22 & 5.45 & 0.321 & 69.87 & 6.25 & 0.340 & 88.01 & 8.48 & 0.411 & 64.73 & 6.17 & 0.342 & 78.05 & 8.27 & 0.413 \\
Nemo 12b       &  & 67.29 & 6.00 & 0.311 & 68.23 & 6.49 & 0.331 & 92.35 & 8.69 & 0.412 & 70.17 & 6.67 & 0.340 & 88.43 & 8.32 & 0.404 \\
Qwen2.5 7b     &  & 68.55 & 6.20 & 0.335 & 70.02 & \textcolor{blue}{6.59} & \textcolor{red}{0.352} & \textcolor{blue}{92.84} & 9.00 &  \textcolor{red}{0.456} & \textcolor{blue}{71.55} & 6.73 & 0.361 & \textcolor{blue}{89.66} & 8.55 & 0.437 \\
Gemma2 9b      &  & 69.16 & 6.36 & 0.334 & 71.65 & 6.53 & 0.347 & 92.69 & \textcolor{blue}{9.07} & \textcolor{blue}{0.443} & 71.03 & 6.89 & \textcolor{red}{0.369} & 87.96 & 8.68 & \textcolor{red}{0.451} \\
Phi4 14b       &  & \textcolor{blue}{73.20} & \textcolor{red}{7.07} & \textcolor{blue}{0.340} & \textcolor{blue}{74.18} & \textcolor{red}{7.32} & \textcolor{blue}{0.349} & \textcolor{red}{92.99} & \textcolor{red}{9.20} & 0.419 & 68.30 & \textcolor{red}{7.28} & 0.354 & 80.04 & \textcolor{red}{8.84} & 0.408 \\
Llama3.3 70b   &  & \textcolor{red}{75.83} & \textcolor{blue}{6.53} & \textcolor{red}{0.346} & \textcolor{red}{75.44} & 6.55 & 0.345 & 92.67 & 8.80 & 0.439 & \textcolor{red}{76.57} & \textcolor{blue}{6.91} & \textcolor{blue}{0.366} & \textcolor{red}{93.45} & \textcolor{blue}{8.82} & \textcolor{blue}{0.446} \\
\bottomrule
\end{tabular}
}
\label{tab:model_scores_by_modality}
\end{table}

The models do not seem to benefit significantly from the RAG information, achieving only a marginal gain compared to the zero-shot settings. This is particularly notable in the top-performing models, while models like Llama2 seem to benefit more from the contextual information, which suggests that this side-information is already present in the larger and newer models.

The results on the open-ended questions align with the accuracy observed in the quiz setting. This consistency confirms that the LLM-as-a-judge is a valuable metric for assessing the answer quality. Regarding the semantic similarity metric, the one reported in Figure \ref{fig:bench_results} represents a similarity score derived from the combination of dense and sparse embeddings generated by BGE-M3. This metric is particularly useful as it incorporates lexical similarity into the evaluation.

\begin{figure}[ht]
    \centering
    \includegraphics[width=0.99\linewidth]{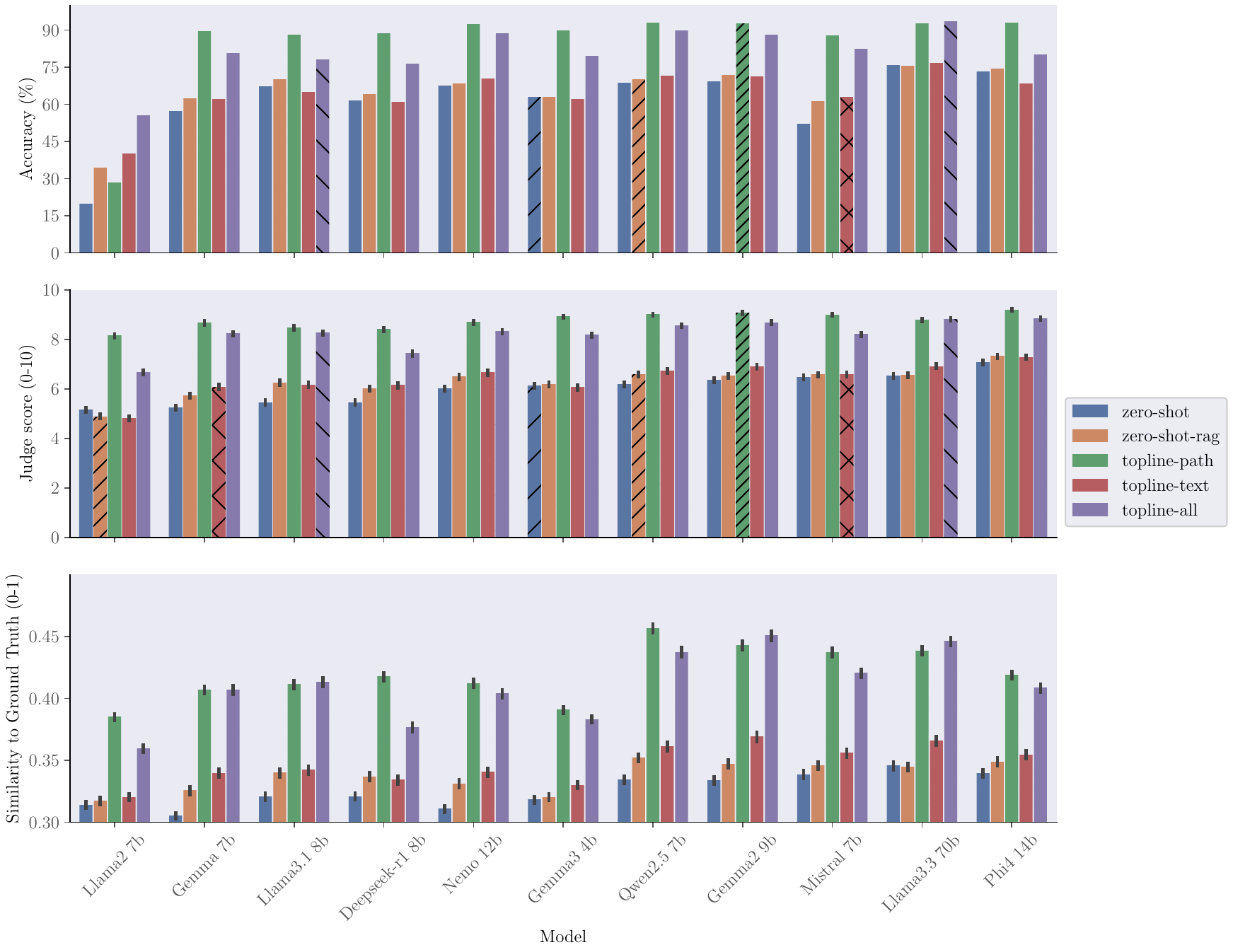}
    \caption{Performances of the models on the HealthBranches dataset. Top: the accuracy of the models on the quiz version in different settings. Middle: the G-eval score in the open-ended answer setting. Bottom: the similarity metric between the predicted answer and ground-truth answer in the open-ended answer setting. To enhance visual clarity, the y-axis is truncated and does not start at zero. This is intended to better highlight the differences between models. Error bars indicate the 95\% confidence interval.}
    \label{fig:bench_results}
\end{figure}

\section{Conclusion}
\label{conclusion}

In this work, we introduced HealthBranches, a benchmark dataset tailored to advance medical question answering through clinically grounded reasoning. By extracting and refining decision pathways from medical literature, we created a resource that not only supports both quiz-style and open-ended formats, but also embeds explicit reasoning chains essential for robust multi-step inference. Our semi-automated pipeline combines structured knowledge extraction, LLM-based generation, and human-in-the-loop refinement, resulting in high-quality Q\&A pairs across 17 clinical domains.

Experimental evaluations across 11 LLMs under zero-shot and RAG conditions confirm that models significantly benefit from structured reasoning paths and underscoring the limitations of relying solely on textual content. Moreover, the LLM-as-a-judge and semantic similarity metrics demonstrate strong alignment with human assessments, reinforcing the dataset’s validity and potential as a rigorous benchmark.

\textbf{Limitations}
Our medical dataset generation pipeline has limitations that should be considered when interpreting its results. While we employ capable language models, potential biases are inevitably introduced by these specific models in the dataset creation process; a challenge that would persist regardless of which model architecture is utilized. The pipeline can propagate any inaccuracies or outdated information present in the source textbooks. Finally, greater attention must be given when working with sensitive medical content, as expert review is necessary. Errors or misrepresentations could affect clinical understanding.

HealthBranches thus provides a step toward the development and assessment of interpretable, safe, and trustworthy LLMs in medical settings. By aligning evaluations with clinical decision-making processes, it offers a valuable resource for both research and medical education.

\section*{Implementation and data availability}
\label{implementation_code}
The dataset is available on \href{https://www.kaggle.com/datasets/earlgrey00/healthbranchesqa/}{Kaggle}. The source code for generating the dataset and experiments: \href{https://anonymous.4open.science/r/HealthBranches-480E/}{anonymous.4open.science/r/HealthBranches-480E}

Details on experimental and hardware setup are available in Appendix~\ref{app:exp_hardware}.

\section*{Acknowledgments}
We thank Chameleon Cloud \cite{cham} for providing the essential computational resources that enabled our experiments and model inference. We thank the medical students and doctors who generously volunteered their time to review the questions. Their feedback was invaluable for the quality and validation of our work.

\bibliographystyle{unsrt}  
\bibliography{references}

\begin{thebibliography}{10}

\bibitem{achiam2023gpt}
Josh Achiam, Steven Adler, Sandhini Agarwal, Lama Ahmad, Ilge Akkaya, Florencia~Leoni Aleman, Diogo Almeida, Janko Altenschmidt, Sam Altman, Shyamal Anadkat, et~al.
\newblock Gpt-4 technical report.
\newblock {\em arXiv preprint arXiv:2303.08774}, 2023.

\bibitem{naveed2023comprehensive}
Humza Naveed, Asad~Ullah Khan, Shi Qiu, Muhammad Saqib, Saeed Anwar, Muhammad Usman, Naveed Akhtar, Nick Barnes, and Ajmal Mian.
\newblock A comprehensive overview of large language models.
\newblock {\em arXiv preprint arXiv:2307.06435}, 2023.

\bibitem{pan2306unifying}
Shirui Pan, Linhao Luo, Yufei Wang, Chen Chen, Jiapu Wang, and Xindong Wu.
\newblock Unifying large language models and knowledge graphs: A roadmap.
\newblock {\em arXiv preprint arXiv:2306.08302}, 2023.

\bibitem{biswas2023role}
Som~S Biswas.
\newblock Role of chat gpt in public health.
\newblock {\em Annals of biomedical engineering}, 51(5):868--869, 2023.

\bibitem{xiong2024benchmarking}
Guangzhi Xiong, Qiao Jin, Zhiyong Lu, and Aidong Zhang.
\newblock Benchmarking retrieval-augmented generation for medicine.
\newblock {\em arXiv preprint arXiv:2402.13178}, 2024.

\bibitem{arslan2024survey}
Muhammad Arslan, Hussam Ghanem, Saba Munawar, and Christophe Cruz.
\newblock A survey on rag with llms.
\newblock {\em Procedia Computer Science}, 246:3781--3790, 2024.

\bibitem{shi2024medical}
Xiaoming Shi, Zeming Liu, Li~Du, Yuxuan Wang, Hongru Wang, Yuhang Guo, Tong Ruan, Jie Xu, and Shaoting Zhang.
\newblock Medical dialogue: A survey of categories, methods, evaluation and challenges.
\newblock {\em arXiv preprint arXiv:2405.10630}, 2024.

\bibitem{carlini2021extracting}
Nicholas Carlini, Florian Tramer, Eric Wallace, Matthew Jagielski, Ariel Herbert-Voss, Katherine Lee, Adam Roberts, Tom Brown, Dawn Song, Ulfar Erlingsson, et~al.
\newblock Extracting training data from large language models.
\newblock In {\em 30th USENIX Security Symposium (USENIX Security 21)}, pages 2633--2650, 2021.

\bibitem{rafailov2024direct}
Rafael Rafailov, Archit Sharma, Eric Mitchell, Christopher~D Manning, Stefano Ermon, and Chelsea Finn.
\newblock Direct preference optimization: Your language model is secretly a reward model.
\newblock {\em Advances in Neural Information Processing Systems}, 36, 2024.

\bibitem{lewis2020retrieval}
Patrick Lewis, Ethan Perez, Aleksandra Piktus, Fabio Petroni, Vladimir Karpukhin, Naman Goyal, Heinrich K{\"u}ttler, Mike Lewis, Wen-tau Yih, Tim Rockt{\"a}schel, et~al.
\newblock Retrieval-augmented generation for knowledge-intensive nlp tasks.
\newblock {\em Advances in Neural Information Processing Systems}, 33:9459--9474, 2020.

\bibitem{liu2023pre}
Pengfei Liu, Weizhe Yuan, Jinlan Fu, Zhengbao Jiang, Hiroaki Hayashi, and Graham Neubig.
\newblock Pre-train, prompt, and predict: A systematic survey of prompting methods in natural language processing.
\newblock {\em ACM Computing Surveys}, 55(9):1--35, 2023.

\bibitem{guu2020retrieval}
Kelvin Guu, Kenton Lee, Zora Tung, Panupong Pasupat, and Mingwei Chang.
\newblock Retrieval augmented language model pre-training.
\newblock In {\em International conference on machine learning}, pages 3929--3938. PMLR, 2020.

\bibitem{izacard2023atlas}
Gautier Izacard, Patrick Lewis, Maria Lomeli, Lucas Hosseini, Fabio Petroni, Timo Schick, Jane Dwivedi-Yu, Armand Joulin, Sebastian Riedel, and Edouard Grave.
\newblock Atlas: Few-shot learning with retrieval augmented language models.
\newblock {\em Journal of Machine Learning Research}, 24(251):1--43, 2023.

\bibitem{huo2023retrieving}
Siqing Huo, Negar Arabzadeh, and Charles~LA Clarke.
\newblock Retrieving supporting evidence for llms generated answers.
\newblock {\em arXiv preprint arXiv:2306.13781}, 2023.

\bibitem{gao2023retrieval}
Yunfan Gao, Yun Xiong, Xinyu Gao, Kangxiang Jia, Jinliu Pan, Yuxi Bi, Yi~Dai, Jiawei Sun, and Haofen Wang.
\newblock Retrieval-augmented generation for large language models: A survey.
\newblock {\em arXiv preprint arXiv:2312.10997}, 2023.

\bibitem{jin2024health}
Mingyu Jin, Qinkai Yu, Dong Shu, Chong Zhang, Lizhou Fan, Wenyue Hua, Suiyuan Zhu, Yanda Meng, Zhenting Wang, Mengnan Du, et~al.
\newblock Health-llm: Personalized retrieval-augmented disease prediction system.
\newblock {\em arXiv preprint arXiv:2402.00746}, 2024.

\bibitem{wang2024healthq}
Ziyu Wang, Hao Li, Di~Huang, and Amir~M Rahmani.
\newblock Healthq: Unveiling questioning capabilities of llm chains in healthcare conversations.
\newblock {\em arXiv preprint arXiv:2409.19487}, 2024.

\bibitem{gilson2024enhancing}
Aidan Gilson, Xuguang Ai, Thilaka Arunachalam, Ziyou Chen, Ki~Xiong Cheong, Amisha Dave, Cameron Duic, Mercy Kibe, Annette Kaminaka, Minali Prasad, et~al.
\newblock Enhancing large language models with domain-specific retrieval augment generation: A case study on long-form consumer health question answering in ophthalmology.
\newblock {\em arXiv preprint arXiv:2409.13902}, 2024.

\bibitem{procko2024graph}
Tyler~Thomas Procko and Omar Ochoa.
\newblock Graph retrieval-augmented generation for large language models: A survey.
\newblock In {\em 2024 Conference on AI, Science, Engineering, and Technology (AIxSET)}, pages 166--169. IEEE, 2024.

\bibitem{jia2024medikal}
Mingyi Jia, Junwen Duan, Yan Song, and Jianxin Wang.
\newblock medikal: Integrating knowledge graphs as assistants of llms for enhanced clinical diagnosis on emrs.
\newblock {\em arXiv preprint arXiv:2406.14326}, 2024.

\bibitem{wu2024medical}
Junde Wu, Jiayuan Zhu, Yunli Qi, Jingkun Chen, Min Xu, Filippo Menolascina, and Vicente Grau.
\newblock Medical graph rag: Towards safe medical large language model via graph retrieval-augmented generation.
\newblock {\em arXiv preprint arXiv:2408.04187}, 2024.

\bibitem{kharitonova2024leveraging}
Ksenia Kharitonova, David P{\'e}rez-Fern{\'a}ndez, Javier Guti{\'e}rrez-Hernando, Asier Guti{\'e}rrez-Fandi{\~n}o, Zoraida Callejas, and David Griol.
\newblock Leveraging retrieval-augmented generation for reliable medical question answering using large language models.
\newblock In {\em International Conference on Hybrid Artificial Intelligence Systems}, pages 141--153. Springer, 2024.

\bibitem{vilares2019head}
David Vilares and Carlos G{\'o}mez-Rodr{\'\i}guez.
\newblock Head-qa: A healthcare dataset for complex reasoning.
\newblock {\em arXiv preprint arXiv:1906.04701}, 2019.

\bibitem{suri2021mediaqa}
Huqun Suri, Qi~Zhang, Wenhua Huo, Yan Liu, and Chunsheng Guan.
\newblock Mediaqa: A question answering dataset on medical dialogues.
\newblock {\em arXiv preprint arXiv:2108.08074}, 2021.

\bibitem{jin2021disease}
Di~Jin, Eileen Pan, Nassim Oufattole, Wei-Hung Weng, Hanyi Fang, and Peter Szolovits.
\newblock What disease does this patient have? a large-scale open domain question answering dataset from medical exams.
\newblock {\em Applied Sciences}, 11(14):6421, 2021.

\bibitem{jin2019pubmedqa}
Qiao Jin, Bhuwan Dhingra, Zhengping Liu, William Cohen, and Xinghua Lu.
\newblock Pubmedqa: A dataset for biomedical research question answering.
\newblock In {\em Proceedings of the 2019 Conference on Empirical Methods in Natural Language Processing and the 9th International Joint Conference on Natural Language Processing (EMNLP-IJCNLP)}, pages 2567--2577, 2019.

\bibitem{khandekar2024medcalc}
Nikhil Khandekar, Qiao Jin, Guangzhi Xiong, Soren Dunn, Serina Applebaum, Zain Anwar, Maame Sarfo-Gyamfi, Conrad Safranek, Abid Anwar, Andrew Zhang, et~al.
\newblock Medcalc-bench: Evaluating large language models for medical calculations.
\newblock {\em Advances in Neural Information Processing Systems}, 37:84730--84745, 2024.

\bibitem{he2020pathvqa}
Xuehai He, Yichen Zhang, Luntian Mou, Eric Xing, and Pengtao Xie.
\newblock Pathvqa: 30000+ questions for medical visual question answering.
\newblock {\em arXiv preprint arXiv:2003.10286}, 2020.

\bibitem{hendrycks2020measuring}
Dan Hendrycks, Collin Burns, Steven Basart, Andy Zou, Mantas Mazeika, Dawn Song, and Jacob Steinhardt.
\newblock Measuring massive multitask language understanding.
\newblock {\em arXiv preprint arXiv:2009.03300}, 2020.

\bibitem{zuo2025medxpertqa}
Yuxin Zuo, Shang Qu, Yifei Li, Zhangren Chen, Xuekai Zhu, Ermo Hua, Kaiyan Zhang, Ning Ding, and Bowen Zhou.
\newblock Medxpertqa: Benchmarking expert-level medical reasoning and understanding.
\newblock {\em arXiv preprint arXiv:2501.18362}, 2025.

\bibitem{park2021knowledge}
Junwoo Park, Youngwoo Cho, Haneol Lee, Jaegul Choo, and Edward Choi.
\newblock Knowledge graph-based question answering with electronic health records.
\newblock In {\em Machine Learning for Healthcare Conference}, pages 36--53. PMLR, 2021.

\bibitem{zhang2024ultramedical}
Kaiyan Zhang, Sihang Zeng, Ermo Hua, Ning Ding, Zhang-Ren Chen, Zhiyuan Ma, Haoxin Li, Ganqu Cui, Biqing Qi, Xuekai Zhu, et~al.
\newblock Ultramedical: Building specialized generalists in biomedicine.
\newblock {\em Advances in Neural Information Processing Systems}, 37:26045--26081, 2024.

\bibitem{sivasubramaniam2024sm3}
Sithursan Sivasubramaniam, Cedric~E Osei-Akoto, Yi~Zhang, Kurt Stockinger, and Jonathan F{\"u}rst.
\newblock Sm3-text-to-query: Synthetic multi-model medical text-to-query benchmark.
\newblock {\em Advances in Neural Information Processing Systems}, 37:88627--88663, 2024.

\bibitem{book1}
Stuart~B Mushlin and Harry~L Greene.
\newblock {\em Decision making in medicine: an algorithmic approach}.
\newblock Elsevier Health Sciences, 2009.

\bibitem{book2}
Mark Harrison and Ala Mohammed.
\newblock {\em Algorithms for emergency medicine}.
\newblock Oxford University Press, 2023.

\bibitem{book3}
Alexander Goldfarb-Rumyantzev.
\newblock {\em Critical Care Medicine: An Algorithmic Approach E-Book}.
\newblock Elsevier Health Sciences, 2021.

\bibitem{gemini}
Gemini Team, Rohan Anil, Sebastian Borgeaud, Jean-Baptiste Alayrac, Jiahui Yu, Radu Soricut, Johan Schalkwyk, Andrew~M Dai, Anja Hauth, Katie Millican, et~al.
\newblock Gemini: a family of highly capable multimodal models.
\newblock {\em arXiv preprint arXiv:2312.11805}, 2023.

\bibitem{llama3}
Aaron Grattafiori, Abhimanyu Dubey, Abhinav Jauhri, Abhinav Pandey, Abhishek Kadian, Ahmad Al-Dahle, Aiesha Letman, Akhil Mathur, Alan Schelten, Alex Vaughan, et~al.
\newblock The llama 3 herd of models.
\newblock {\em arXiv preprint arXiv:2407.21783}, 2024.

\bibitem{ollama}
https://ollama.com/.

\bibitem{jiang2024identifying}
Fengqing Jiang.
\newblock Identifying and mitigating vulnerabilities in llm-integrated applications.
\newblock Master's thesis, University of Washington, 2024.

\bibitem{touvron2023llama}
Hugo Touvron, Louis Martin, Kevin Stone, Peter Albert, Amjad Almahairi, Yasmine Babaei, Nikolay Bashlykov, Soumya Batra, Prajjwal Bhargava, Shruti Bhosale, et~al.
\newblock Llama 2: Open foundation and fine-tuned chat models.
\newblock {\em arXiv preprint arXiv:2307.09288}, 2023.

\bibitem{grattafiori2024llama}
Aaron Grattafiori, Abhimanyu Dubey, Abhinav Jauhri, Abhinav Pandey, Abhishek Kadian, Ahmad Al-Dahle, Aiesha Letman, Akhil Mathur, Alan Schelten, Alex Vaughan, et~al.
\newblock The llama 3 herd of models.
\newblock {\em arXiv preprint arXiv:2407.21783}, 2024.

\bibitem{team2024gemma}
Gemma Team, Thomas Mesnard, Cassidy Hardin, Robert Dadashi, Surya Bhupatiraju, Shreya Pathak, Laurent Sifre, Morgane Rivi{\`e}re, Mihir~Sanjay Kale, Juliette Love, et~al.
\newblock Gemma: Open models based on gemini research and technology.
\newblock {\em arXiv preprint arXiv:2403.08295}, 2024.

\bibitem{team2024gemma2}
Gemma Team, Morgane Riviere, Shreya Pathak, Pier~Giuseppe Sessa, Cassidy Hardin, Surya Bhupatiraju, L{\'e}onard Hussenot, Thomas Mesnard, Bobak Shahriari, Alexandre Ram{\'e}, et~al.
\newblock Gemma 2: Improving open language models at a practical size.
\newblock {\em arXiv preprint arXiv:2408.00118}, 2024.

\bibitem{team2025gemma}
Gemma Team, Aishwarya Kamath, Johan Ferret, Shreya Pathak, Nino Vieillard, Ramona Merhej, Sarah Perrin, Tatiana Matejovicova, Alexandre Ram{\'e}, Morgane Rivi{\`e}re, et~al.
\newblock Gemma 3 technical report.
\newblock {\em arXiv preprint arXiv:2503.19786}, 2025.

\bibitem{yang2024qwen2}
An~Yang, Baosong Yang, Beichen Zhang, Binyuan Hui, Bo~Zheng, Bowen Yu, Chengyuan Li, Dayiheng Liu, Fei Huang, Haoran Wei, et~al.
\newblock Qwen2. 5 technical report.
\newblock {\em arXiv preprint arXiv:2412.15115}, 2024.

\bibitem{guo2025deepseek}
Daya Guo, Dejian Yang, Haowei Zhang, Junxiao Song, Ruoyu Zhang, Runxin Xu, Qihao Zhu, Shirong Ma, Peiyi Wang, Xiao Bi, et~al.
\newblock Deepseek-r1: Incentivizing reasoning capability in llms via reinforcement learning.
\newblock {\em arXiv preprint arXiv:2501.12948}, 2025.

\bibitem{abdin2024phi}
Marah Abdin, Jyoti Aneja, Harkirat Behl, S{\'e}bastien Bubeck, Ronen Eldan, Suriya Gunasekar, Michael Harrison, Russell~J Hewett, Mojan Javaheripi, Piero Kauffmann, et~al.
\newblock Phi-4 technical report.
\newblock {\em arXiv preprint arXiv:2412.08905}, 2024.

\bibitem{toge}
https://www.together.ai/.

\bibitem{g-eval}
Yang Liu, Dan Iter, Yichong Xu, Shuohang Wang, Ruochen Xu, and Chenguang Zhu.
\newblock {G}-eval: {NLG} evaluation using gpt-4 with better human alignment.
\newblock In Houda Bouamor, Juan Pino, and Kalika Bali, editors, {\em Proceedings of the 2023 Conference on Empirical Methods in Natural Language Processing}, pages 2511--2522, Singapore, December 2023. Association for Computational Linguistics.

\bibitem{bge}
Jianlv Chen, Shitao Xiao, Peitian Zhang, Kun Luo, Defu Lian, and Zheng Liu.
\newblock Bge m3-embedding: Multi-lingual, multi-functionality, multi-granularity text embeddings through self-knowledge distillation.
\newblock {\em arXiv preprint arXiv:2402.03216}, 2024.

\bibitem{cham}
Kate Keahey, Jason Anderson, Zhuo Zhen, Pierre Riteau, Paul Ruth, Dan Stanzione, Mert Cevik, Jacob Colleran, Haryadi~S Gunawi, Cody Hammock, et~al.
\newblock Lessons learned from the chameleon testbed.
\newblock In {\em 2020 USENIX annual technical conference (USENIX ATC 20)}, pages 219--233, 2020.

\end{thebibliography}


\appendix
\section{Supplementary Material} 
\label{app:supplementary}

\subsection{Dataset details}
\label{app:dataset_details}
In this section, we present the prompts used for data extraction using Gemini. 

\begin{promptbox}[title = Prompt used in the experiments to extract the graphical stream]

For each pdf file that i send you will respond in the following way:
\begin{enumerate}
    \item  Put the content of the decision tree/graph in a valid JSON format. It has to be structured as a decision tree. If the branch of a decision is unclear or missing, get it from the text. The JSON must contain these keys:
    \begin{enumerate}[label=\alph*)]
        \item node: containing the node name
        \item content: a brief description of the node
        \item children: a list of children node (can be optional for leaf nodes)
        \item make a root node with the name of the condition
    \end{enumerate}
\item Make sure that the JSON is valid and well structured. Make sure that you don't get 'Invalid control character at line' errors.
\end{enumerate}
\end{promptbox}

The system prompt used to polish raw text extracted from the PDF (textual stream) is:

\begin{promptbox}[title = Prompt used to refine raw text extracted from a PDF]
The uploaded PDF contains information and decisions to be made about a specific disease/condition.

Each PDF consists of (sorted from the top of the first page): a title, the authors, a brief description, observations (begin with a capital letter followed by a period), a decision graph, a continuation of an observation if it did not fit on the previous page, references.

In some documents there may be tables, ignore them.
\end{promptbox}

While the single message with the single condition attached (PDF file):

\begin{promptbox}[title = Prompt to extract the condition from the PDF file]
Analyze the PDF provided and extract the brief description and observations.
The title, authors, decision graph and references should NOT be extracted.
There must be ONLY text contained in the provided pdf in the output.
\end{promptbox}

Additionally, we provided an example of path refinement using a real medical case.

\begin{promptbox}[title = System Prompt (path refinement) ]

\texttt{Given a sequence of reasoning and associated condition, you have to refine it to be uniform with medical terminology.
\newline
 The meaning of the sequence must NOT change (at most, you can remove superfluous information or sanitize the text). If the reasoning step is poorly explained or ambiguous, refine it using the context and medical knowledge.
\newline
The answer must be a sequence of reasoning with -> indicating the transition between one step and the next. 
\newline
Do not add any more text or reasoning in your answer, just the sequence.
}
\end{promptbox}

 \begin{tcolorbox}[
   enhanced,
   colback=white,
   colframe=green!50!black,
   sharp corners,
   boxrule=0.5pt,
   toptitle=0.3mm,
   bottomtitle=0.3mm,
   title=\textbf{Pleural Effusion Decision Paths before and after refinement process},
   fonttitle=\bfseries,
   fontupper=\ttfamily\fontsize{9.5pt}{11.5pt}\selectfont,
   coltitle=white]
 \begin{minipage}[t]{0.47\textwidth}
   \textbf{Old Path}
   \begin{enumerate}[leftmargin=*, label=\arabic*.,  itemsep=0pt, parsep=0pt, topsep=0pt]
     \item Imaging: PA \& lateral CXR; consider decubitus film; consider ultrasound
     \item Suspect CHF or viral pleurisy
     \item No or atypical progression
     \item Thoracentesis
     \item Light’s criteria: TP\textsubscript{f}/TP\textsubscript{s} > 0.5. LDH\textsubscript{f}/LDH\textsubscript{s} > 0.6. LDH\textsubscript{pf} > 2/3 upper limit of normal.
     \item Exudate
     \item Cell count \& differential
     \item Mononuclear cell predominant. Viral infection. Chronic causes
   \end{enumerate}
 \end{minipage}
 \hfill
 \begin{minipage}[t]{0.47\textwidth}
   \textbf{New Path}
   \begin{enumerate}[leftmargin=*, label=\arabic*., itemsep=0pt, parsep=0pt, topsep=0pt]
     \item Obtain radiological images: PA \& lateral CXR; consider decubitus film or ultrasound
     \item Suspect CHF or viral pleurisy
     \item No or atypical progression
     \item Perform thoracentesis
     \item Apply Light’s criteria: TP\textsubscript{f}/TP\textsubscript{s} > 0.5.LDH\textsubscript{f}/LDH\textsubscript{s} > 0.6. LDH\textsubscript{pf} > 2/3 upper limit of normal.
     \item Exudative effusion identified
     \item Perform cell count \& differential
     \item Mononuclear cell predominance
     \item Consider viral infection or chronic etiologies
   \end{enumerate}
 \end{minipage}
 \end{tcolorbox}

\begin{promptbox}[title = System Prompt (question generation) ]
\texttt{Given a sequence of reasoning and a text related to it about how to treat a symptom/condition, generate:
\newline
1. a question reflecting the reasoning of the sequence provided. The question must include a clinical case, e.g: "A 67-year-old man is brought to the physician because of increasing forgetfulness, unsteadiness, and falls over the past year..."
\newline
2. A set of 5 possible answers (A,B,C,D,E). Should not be too long and should also reflect the sequence of reasoning. One of them must be the correct answer. The other answers need not be correct for the generated question but must be related to the topic of the question.
\newline
The sequence does NOT have to be explicit in both question and answers!
\newline
The correct option must be the same as the answer.
\newline
The output should be structured in the following format: ["question", "answer", "['Option A', 'Option B', 'Option C', 'Option D', 'Option E']", "letter of correct option"]
\newline
Do not generate any additional texts.
}
\end{promptbox}

\begin{promptbox}[title = Generation question prompt ]
\texttt{The reasoning sequence is as follows: \{path\}, the context associated is: \{text\} and the symptom/condition to be treated is: \{cond\}.}
\end{promptbox}

\subsection{Extract option rule set}
\label{app:rules}
Below is the algorithm used for extracting an option from a string.

\begin{algorithm}
  \caption{ExtractOption(answer)}
  \begin{algorithmic}[1]
    \Procedure{ExtractOption}{answer}
      \State ans $\gets$ \textsc{Trim}(answer)
      \ForAll{rule $R_i$ in $\{R1,\dots,R8\}$}
        \If{ans matches $R_i$.pattern}
          \State ch $\gets$ uppercase(captured letter)
          \State \Return ch
        \EndIf
      \EndFor
      \State \Return None
    \EndProcedure
  \end{algorithmic}
\end{algorithm}

\vspace{1ex}
\noindent\textbf{Rule definitions:}
\begin{itemize}
  \item[R1:] \verb|^[A-Ea-e]$|  
    — exact single letter  
  \item[R2:] \verb|^([A-Ea-e])\s|  
    — letter + space at start  
  \item[R3:] \verb|^([A-Ea-e])\.|  
    — letter + period at start  
  \item[R4:] \verb|\(([A-Ea-e])\)| or \verb|\s([A-Ea-e])\)|  
    — letter in parentheses or before “)”  
  \item[R5:] \verb|^[A-E][A-Z]|  
    — leading uppercase + noise  
  \item[R6:] \verb|^[A-Ea-e][^A-Za-z]|  
    — letter + non‑letter at start  
  \item[R7:] any of
    \begin{itemize}
      \item \verb|the correct answer is[:\s]*\(?([A-Ea-e])\)?|
      \item \verb|the answer is[:\s]*\(?([A-Ea-e])\)?|
      \item \verb|option *\(?([A-Ea-e])\)?|
      \item \verb|answer: *\(?([A-Ea-e])\)?|
    \end{itemize}
  \item[R8:] \verb|[:\.,]\s*([A-Ea-e])$|  
    — trailing letter after punctuation  
\end{itemize}

\subsection{Q\&A validation} \label{app:qa_valid}
This section reports the prompt used to validate the questions that were incorrectly answered by both Llama3.3 (70 B) and Llama3.1 (405 B), along with the corresponding answers and path.
In addition, the template used to submit the Q\&A is presented using an example with the matching answer given by ChatGPT applied with web search and reasoning capabilities. 

\begin{promptbox}[title = Prompt submitted to ChatGPT]
\texttt{You will now act as a reviewer for some questions and answers to a quiz on the medical domain. I will give you a question, some possible answers, the supposed correct answer and a reasoning path that can help to respond correctly which is structured exactly like a path with arrows separating each step of the path.Your task is to check if the question makes sense, if the reported correct option is the correct option, if other options may be also correct and if the reasoning path is correct. You must check the soundness of the quiz using existing medical literature and clinical guidelines. If there are any issues report them in an order manner and suggest which modification should be done to the quiz in order to make it right.The modification can be of 4 types: 
\begin{enumerate}[start=0]
    \item The question is regarded as complex or ambiguous to answer.
    \item Change the correct answer (i.e. change the correct option to another or adding/correcting an option).
    \item Change the question so that the resulting option is correct (i.e. if the question has some imprecise or missing information).
    \item If these changes are affecting the reasoning path change it accordingly but keep the same schematic/path style (make sure to keep the same action like structure using ->).
\end{enumerate}
 After evaluating all of this, suggest the simplest modification between 0, 1, 2, or 3 to make the quiz sound. Give a brief motivation about it. Finally, output the quiz in the same template I sent it to you, but corrected with your best suggestion.}
\end{promptbox}

\begin{promptbox}[title = Example of Q\&A using our template]
\texttt{Q\&A template:}
\begin{itemize}
  \item Question: \textit{A 34-year-old male is newly diagnosed with HIV. What is the correct order of actions to take after the diagnosis?}\\
  \item Options: \textit{['Prescribe prophylactic antibiotics.', 'Refer the patient for immediate psychological counseling.', 'Perform a lumbar puncture.', 'Administer tetanus, diphtheria, and acellular pertussis boosters.', 'Initiate antiretroviral therapy immediately.']}\\
  \item  Correct option: \textit{Administer tetanus, diphtheria, and acellular pertussis boosters.} \\
  \item  Reasoning path: \textit{ Initial steps for newly diagnosed HIV-positive patients -> Immunizations -> Tetanus, diphtheria, and acellular pertussis boosters.} \\
\end{itemize}

\end{promptbox}

\textit{Reasoned about the consistency of the medical quiz for 27 seconds.}

\begin{promptbox}[title = ChatGPT  answer]
Below is the evaluation and recommendation:

\textbf{ISSUES IDENTIFIED:}
\begin{enumerate}
    \item \textbf{Incorrect Correct Option:}
    \begin{itemize}
        \item Observation: Current clinical guidelines for newly diagnosed HIV patients prioritize the initiation of antiretroviral therapy (ART) immediately. Immunizations (such as tetanus, diphtheria, and acellular pertussis boosters) are important, but they do not supersede the need for immediate ART initiation.
        \item Literature/Guidelines: The U.S. Department of Health and Human Services (DHHS) and WHO both recommend that ART should be started as soon as possible after diagnosis.
        \item Recommendation: Change the correct answer to “Initiate antiretroviral therapy immediately.”
    \end{itemize}
    \item \textbf{Question Wording Ambiguity:}
    \begin{itemize}
     \item Observation: The phrasing “order of actions” implies a sequential management plan. However, the provided options mix interventions that occur at different stages.
     \item Recommendation: To avoid ambiguity, modify the question so that it asks for the “most immediate” or “priority” intervention after an HIV diagnosis.
      \end{itemize}
   \item \textbf{Reasoning Path Issues:}
   \begin{itemize}
       \item Observation: The current reasoning path directs the reader toward immunizations rather than immediate ART initiation.
       \item Recommendation: Update the reasoning path to emphasize that immediate ART is the priority, with immunizations coming later in the comprehensive care plan.
\end{itemize}
\end{enumerate}
\textbf{SIMPLEST MODIFICATION CHOSEN:}
\begin{itemize}
    \item \textbf{Modification Type 1:} Change the correct answer (and adjust the reasoning path accordingly).This is the simplest modification because it only requires updating the correct answer and its supporting reasoning without having to rework the entire quiz structure.
\end{itemize}
\end{promptbox}

As an overview of the possible corrections made in this phase, Table \ref{tab:horizontal_questions} illustrates four examples of corrections, one for each category (0, 1, 2, and 3), following the established protocol.

\begin{sidewaystable}
\centering
\scriptsize
\begin{tabular}{|p{3.5cm}|p{3.5cm}|p{5.5cm}|p{3.5cm}|p{2.5cm}|p{3.5cm}|}
\hline
\textbf{Type} & \textbf{Question} & \textbf{Options} & \textbf{Path} & \textbf{Modifications} & \textbf{Motivation} \\ \hline

\textbf{0 - Complex or ambiguous question} &
A 54-year-old woman is evaluated for a diastolic murmur. The murmur is early diastolic with a decrescendo configuration, heard \hlightpurple{best at the left lower sternal border, and increases in intensity with inspiration.} Which of the following is the most likely underlying cause? &
\begin{itemize}[leftmargin=*]
\item[A.] Aortic regurgitation.
\item[B.] \hlightpurple{\textit{Pulmonic regurgitation.}}
\item[C.] Ventricular septal defect.
\item[D.] Tricuspid stenosis.
\item[E.] Mitral stenosis.
\end{itemize} &
Patient with diastolic murmur -> Early diastolic decrescendo sound -> Right-sided murmur (increases with inspiration) -> Consistent with pulmonic regurgitation &
None &
A diastolic decrescendo murmur that intensifies with inspiration complicates the differentiation between pulmonic and tricuspid regurgitation, thereby contributing to the ambiguity. \\ \hline

\textbf{1 - Change the correct answer} &
A 58-year-old male presents with a solitary pulmonary nodule (SPN) discovered incidentally on a chest CT scan. The nodule is less than 3 cm in diameter, completely surrounded by lung parenchyma, and without associated lymphadenopathy. Previous chest imaging from 3 years ago shows the nodule to be stable. What is the most appropriate next step in management? &
\begin{itemize}[leftmargin=*]
\item[A.] CT-guided transthoracic needle aspiration.
\item[B.] Repeat CT chest at 3, 6, 9, 12, and 24 months.
\item[C.] No further follow-up is necessary.
\item[D.] Positron emission tomography (PET) scan.
\item[E.] \hlightorange{\textit{Thoracic surgery referral for VATS/thoracotomy.}}
\end{itemize} &
Patient with solitary pulmonary nodule -> Lesion <3 cm surrounded by lung parenchyma in absence of other abnormalities -> Previous chest imaging available, demonstrating stability for >2 years? -> Thoracic surgery referral for VATS/thoracotomy solitary pulmonary nodule &
\hlightgreen{Correct answer changed from E to C: “No further follow-up is necessary.”} &
The stability of the nodule over 3 years suggests a benign nature, consistent with guidelines that recommend no further follow-up for such lesions. \\ \hline

\textbf{2 - Change the question} &
A 55-year-old female presents with difficulty swallowing solids. \hlightorange{Initial assessment, including manometry, shows normal esophageal motility.} What is the MOST appropriate next step in evaluating her dysphagia? &
\begin{itemize}[leftmargin=*]
\item[A.] \textit{Perform barium esophagography to assess for extrinsic compression.}
\item[B.] Esophageal manometry to assess esophageal motility.
\item[C.] Referral to speech pathologist for swallowing therapy.
\item[D.] Trial of proton pump inhibitors to treat potential GERD.
\item[E.] Esophagogastroduodenoscopy (EGD) to visualize the esophageal mucosa.
\end{itemize} &
Patient presents with difficulty swallowing solids -> Conduct initial evaluation with EGD and manometry -> Normal mucosal and motility findings -> Perform barium esophagography -> Identify extrinsic compression as the cause &
A 55-year-old female presents with difficulty swallowing solids. \hlightgreen{Initial evaluation including EGD and manometry reveals normal esophageal mucosa and motility.} What is the MOST appropriate next step in evaluating her dysphagia for suspected extrinsic compression? &
Non-diagnostic endoscopic and motility findings raise suspicion for extrinsic compression, supporting the need for barium study.  \\ \hline

\textbf{3 - Change the path} &
A 30-year-old patient presents with symptoms of depression, including low mood, fatigue, and difficulty concentrating, which \hlightorange{started a few weeks after starting a new medication for hypertension}. The patient denies any prior history of depression or substance abuse. What is the most appropriate next step in evaluating the patient's depression? &
\begin{itemize}[leftmargin=*]
\item[A.] Immediately start the patient on a selective serotonin reuptake inhibitor (SSRI).
\item[B.] \hlightorange{\textit{Discontinue antihypertensive medication and monitor for resolution of depressive symptoms.}}
\item[C.] Obtain a detailed medical and psychiatric history, including a review of all current medications and substances used, and perform a physical and mental status examination.
\item[D.] Refer the patient for cognitive behavioral therapy.
\item[E.] Order a dexamethasone suppression test to assess for major depression.
\end{itemize} &
\hlightorange{Medical and psychiatric history. Medication/substance history. Physical examination. Mental status examination ->  Consider: Organic mood disorder, depressed. Major depression ->  Specific organic factor not present -> Adjustment disorder with depressed mood -> Specific diagnostic tests ->  Eliminate drug} &
\hlightgreen{New antihypertensive started -> Onset of depressive symptoms soon after -> High suspicion for medication-induced depression -> Discontinue the offending agent and monitor for resolution} &
The onset of symptoms following the initiation of antihypertensive therapy aligns with the DSM-5 criteria for medication-induced depression, warranting discontinuation of the medication and careful observation prior to considering further treatment options.\\ \hline
\end{tabular}
\caption{Examples of correction types detected by ChatGPT. The correct answer choices are indicated in \textit{italics}. Elements that prompt modifications are highlighted in \hlightorange{orange}, the applied corrections are shown in \hlightgreen{green}, and content identified as complex or ambiguous, whether in the question stem or answer options, is marked in \hlightpurple{purple}.}
\label{tab:horizontal_questions}
\end{sidewaystable}

\subsection{Quiz Inference} 
\label{app:supp_quiz}
We present the prompts used during the evaluation phase of the quiz dataset, employing various models and setups.
\begin{promptbox}[title=Zero shot prompt]
The following is a multiple-choice question about how to manage a patient affected by \{condition\}. 

Reply ONLY with the letter (A,B,C,D,E) of the answer you think is CORRECT, without any additional text!

Question: \{question\}

Options:
\begin{itemize}
    \item A. \{option1\}
    \item B. \{option2\}
    \item C. \{option3\}
    \item D. \{option4\}
    \item E. \{option5\}
\end{itemize}
    
The answer MUST be a single letter.

Correct option:
\end{promptbox}
\begin{promptbox}[title=Zero shot RAG prompt]
The following is a multiple-choice question about how to manage a patient affected by \{condition\}. 

Reply ONLY with the letter (A,B,C,D,E) of the answer you think is CORRECT, without any additional text!

Answer the question using the provided context.

Context: \{context\}

Question: \{question\}

Options:
\begin{itemize}
    \item A. \{option1\}
    \item B. \{option2\}
    \item C. \{option3\}
    \item D. \{option4\}
    \item E. \{option5\}
\end{itemize}
    
The answer MUST be a single letter.

Correct option:
\end{promptbox}
\begin{promptbox}[title=Zero shot Topline (path only) prompt]
The following is a multiple-choice question about how to manage a patient affected by \{condition\}. 

Reply ONLY with the letter (A,B,C,D,E) of the answer you think is CORRECT, without any additional text!

The choice of the correct answer should be based on the following sequence of reasoning: \{path\}.

Question: \{question\}

\begin{itemize}
    \item A. \{option1\}
    \item B. \{option2\}
    \item C. \{option3\}
    \item D. \{option4\}
    \item E. \{option5\}
\end{itemize}
    
The answer MUST be a single letter.

Correct option:
\end{promptbox}
\begin{promptbox}[title=System instruction for deepseek]
 Answer the question directly with the letter of the correct option only (e.g., A, B, C, D, or E).
 
Do NOT include any internal reasoning or chain-of-thought in your response.
\end{promptbox}

\subsection{Open answer inference} 
We present the prompts used during the evaluation phase of the open-ended dataset, employing various models and setups.
\label{app:supp_open}
\begin{promptbox}[title=Zero shot prompt]
Answer the question about how to manage a patient affected by \{condition\}. The answer must:

\begin{itemize}
    \item no longer than 100/150 characters;
    \item must contain a decision made on the decision-making process of a clinical case described in the question.
\end{itemize}

Question: \{question\}

Answer:
\end{promptbox}
\begin{promptbox}[title=Zero shot RAG prompt]
Answer the question about how to manage a patient affected by \{condition\}. The answer must:
\begin{itemize}
    \item no longer than 100/150 characters;
    \item must contain a decision made on the decision-making process of a clinical case described in the question.
\end{itemize}

Answer the question using the provided context.

Context: \{context\}

Question: \{question\}

Answer:
\end{promptbox}
\begin{promptbox}[title=Zero shot topline (path only) prompt]
Answer the question about how to manage a patient affected by \{condition\}. The answer must:

\begin{itemize}
    \item no longer than 100/150 characters;
    \item must contain a decision made on the decision-making process of a clinical case described in the question.
\end{itemize}

The answer must be based on the following reasoning path: \{path\}.

Question: \{question\}

Answer:
\end{promptbox}
\begin{promptbox}[title=System instruction for deepseek]
 Answer the question directly.
 
Do NOT include any internal reasoning or chain-of-thought in your response.
\end{promptbox}

\subsection{Inference times details}
\label{app:inf_times}
The table below presents the average inference times required to generate a single response in both the quiz and open-ended answer settings.

\begin{table}[ht!]
\centering
\begin{tabular}{lrr}
\toprule
Model & Avg. Quiz & Avg. Open \\
\midrule
gemma3:4b         & 0.528 & 0.738 \\
mistral:7b        & 0.440 & 0.906 \\
gemma:7b          & 0.449 & 0.834 \\
qwen2.5:7b        & 0.413 & 0.691 \\
llama2:7b         & 0.489 & 1.567 \\
llama3.1:8b       & 0.547 & 1.012 \\
deepseek-r1:8b    & 7.233 & 6.000 \\
gemma2:9b         & 0.572 & 0.935 \\
mistral-nemo:12b  & 0.574 & 0.746 \\
phi4:14b          & 1.429 & 2.240 \\
\bottomrule
\end{tabular}
\end{table}

\subsection{LLM-as-a-judge implementation} 
\label{app:supp_judge}
Prompt used for the evaluation phase with the LLM-as-a-judge.

\begin{promptbox}[title=LLM-as-a-judge prompt]
*** TASK:
Based on the following task description and evaluation criteria,
generate a detailed Chain of Thought (CoT) that outlines the necessary Evaluation Steps
to assess the solution. The CoT should clarify the reasoning process for each step of evaluation.

*** INPUT:

TASK INTRODUCTION:
You are an evaluator for Q\&A medical tasks. You will evaluate the quality of answers to medical questions. You will have some ground truth to help you evaluate.

You will be given 4 inputs:
\begin{itemize}

    \item The question
    \item The ground truth answer
    \item The reasoning path that entails the answer
    \item The given answer which needs to be evaluated
\end{itemize}

The inputs will be given in csv format as a batch of 3 samples). The columns will contain:
\begin{itemize}
    \item question
    \item reasoning path
    \item real answer
    \item predicted answer
\end{itemize}

The output must be a csv formatted including ALL the examples given in the input. But only of a column indicating the score

*** EVALUATION CRITERIA:

The evaluation criteria must take into account how much the given answer (input 4) is adherent to the reasoning path and to the ground truth answer.

The evaluation MUST not take into account the "style" of the answer but only the content.
The score must be an integer between 0 to 10.

FINAL SCORE:

IF THE USER'S SCALE IS DIFFERENT FROM THE 0 TO 10 RANGE, RECALCULATE THE VALUE USING THIS SCALE.

SCORE VALUE MUST BE AN INTEGER.
\end{promptbox}

\subsection{Experimental and hardware setup}
\label{app:exp_hardware}
All experiments were run on a high-performance node equipped with an NVIDIA Quadro RTX 6000 GPU, an Intel Xeon Gold CPU, 192 GB RAM and Ubuntu Linux. Inference was performed using the Ollama framework (v0.6.3), with CUDA and cuDNN support. Statistics on the inference times are available in Appendix \ref{app:inf_times}. Two quantization schemes were used: \texttt{Q\_0}: 8-bit weight representation, no further compression (Gemma1/2, Mistral, Llama2, Mistral-nemo). \texttt{Q4\_K\_M}: 4-bit k-means quantization (Gemma3, Qwen2.5, Deepseek-r1, Llama3.1, Phi4).

The experiments were completed on April 12th 2025; therefore, models released after this date are not included in the benchmark evaluation.

\subsection{Medical Validation Protocol}
\label{app:med_val}

This section describes the structured protocol used by medical professionals to review and validate the Q\&A content in our dataset.
The following protocol was administered to students, medical specialists, and doctors using a simple and user-friendly web app to evaluate a sample of Q\&A.

\begin{enumerate}
  \item \textbf{Purpose.}  
    To provide a structured framework for medical professionals to assess and ensure the accuracy, clarity, dependability, and identification of potential misinformation or factual inaccuracies in the Q\&A content provided to patients and the general public.
  
  \item \textbf{Scope.}  
    Applicable to all medical professionals engaged in the evaluation process (physicians, specialists, medical researchers). Focuses on verifying:
    \begin{itemize}
      \item Content accuracy  
      \item Adherence to current medical guidelines  
      \item Suitability for patient education  
    \end{itemize}
  
  \item \textbf{Reviewer Qualifications.}  
    Individuals must meet at least one of the following criteria:
    \begin{itemize}
      \item Medical Student  
      \item Doctor / Physician  
      \item Medical Specialist  
    \end{itemize}
  
  \item \textbf{Review Process.}
    \begin{enumerate}[label=4.\arabic*]
      \item \textbf{Selection of Q\&A Content.}  
        A curated set of Q\&A instances is chosen to span a broad range of medical topics.
      
      \item \textbf{Evaluation Parameters and Scoring.}  
        Each Q\&A entry is scored on a 1–5 scale (1 = Poor, 5 = Excellent) according to:
        \begin{itemize}
          \item \emph{Question Medical Accuracy:}  
            Alignment with current medical knowledge and reliable sources.
          \item \emph{Answer Medical Accuracy:}  
            Correctness of the diagnosis or explanation. If incorrect, the reviewer selects the appropriate answer.
          \item \emph{Path Medical Accuracy:}  
            Correctness and completeness of the management steps outlined.
        \end{itemize}
        Total score ranges from 3 to 15. Entries scoring below a predefined threshold (i.e., 9/15) are flagged as incorrect.
      
      \item \textbf{Providing Feedback and Documentation.}  
        \begin{itemize}
          \item Entries below threshold of 3/5: reviewers submit detailed feedback highlighting strengths and areas for revision.  
        \end{itemize}
      
      \item \textbf{Structured Review Form.}  
        A web-app form is used to collect expert reviews:
        \begin{itemize}
          \item Each Q\&A instance is displayed with fields for scores and free‐text comments.  
          \item The form auto‐calculates total scores and flags entries below threshold.  
          \item Aggregated results are exported for trend analysis and identification of common issues.
        \end{itemize}
    \end{enumerate}
  
  \item \textbf{Implementation of Revisions.}
    \begin{itemize}
      \item The Q\&A development team reviews all feedback.  
      \item Clarifications or further expert input are obtained as needed.  
      \item A final verification pass is performed before publishing revisions.
    \end{itemize}
  
  \item \textbf{Ongoing Review and Quality Assurance.}
    \begin{itemize}
      \item Periodic re‐assessment of Q\&A content to ensure continued accuracy and relevance.  
      \item Analysis of reviewer feedback trends to guide future improvements.  
      \item Protocol updates to incorporate advances in medical science and best practices.
    \end{itemize}
  
  \item \textbf{Confidentiality and Ethical Considerations.}
    \begin{itemize}
      \item Reviewers must adhere to confidentiality agreements regarding unpublished material.  
      \item All activities must comply with ethical standards and relevant regulatory requirements for medical information dissemination.
    \end{itemize}
\end{enumerate}

\subsection{Web App Review Examples}
\label{app:med_spec}
To illustrate the detailed corrections for Q\&A that received at least one score of 3 or lower, two examples of the feedback provided by clinicians involved in this phase are displayed in Figure \ref{fig:med_spec_rev}. These suggestions incorporate field-specific details to enhance the Q\&A.
\begin{figure}[ht]
    \centering
    \includegraphics[width=\linewidth]{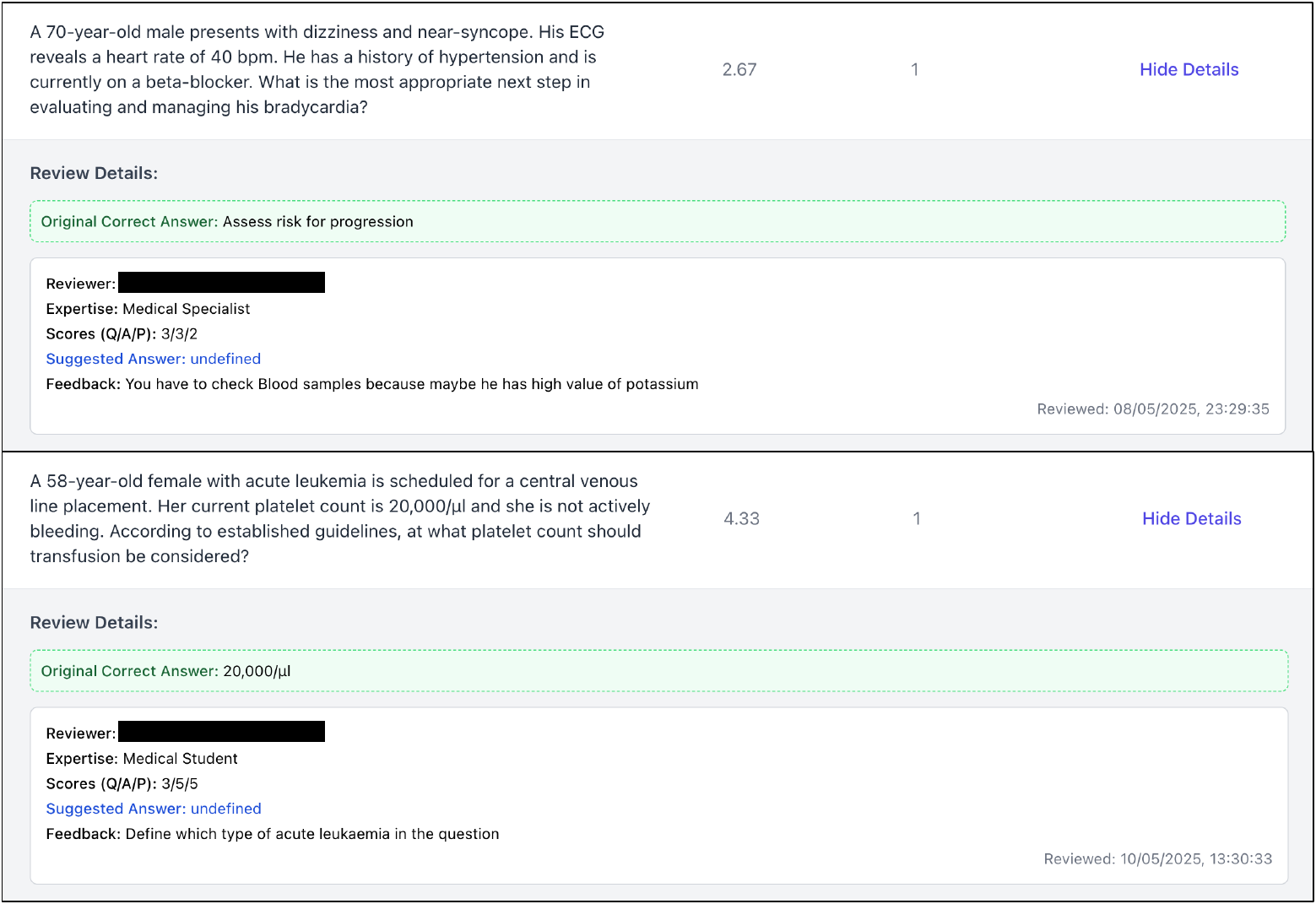}
    \caption{Examples of reviews conducted by a medical specialist and a medical student receiving at least one score of 3 or lower.}
    \label{fig:med_spec_rev}
\end{figure}

\end{document}